# A Deep Reinforcement Learning Model for Predictive Maintenance Planning of Road Assets: Integrating LCA and LCCA

Moein Latifi[1†] and Fateme Golivand Darvishvand[1†]

Omid Khandel[2], Mobin Latifi Nowsoud[3]

[1] AmirKabir University of Technology{m.lattifi,f.g.darvishvand}@aut.ac.ir

[2] Oklahoma State University, omidk@okstate.edu

[3] Islamic Azad University, mobin.latifi@gmail.com

[†] These authors contributed equally to this work

**Abstract**. One of the main challenges in maintenance and rehabilitation (M&R) practices is to determine the type and timing of the maintenance actions. Determining the right maintenance time and type can result in saving a significant amount of budget. This paper proposes a framework using Reinforcement Learning (RL) to determine the optimum type and timing of M&R practices based on the Long-Term Pavement Performance (LTPP) data. A predictive Deep Neural Network (DNN) model is developed in the proposed algorithm, serving as the RL algorithm's Environment. For the policy estimation of the RL model, the efficacy of DQN and PPO models are evaluated. International Roughness Index (IRI) and Rutting Depth (RD) are used as pavement performance indicators. Economic and environmental impacts of M&R treatments are considered in reward (i.e., cost-effectiveness) calculations. Costs and environmental impacts are evaluated using paLATE 2.0 software. In order to verify model convergence, the sensitivity of the model with respect to different traffic volumes is tested. To cover both warm and wet climatic conditions, the proposed framework is applied to a case study located in the state of Texas; however, it is equally applicable to various transportation networks across the globe. The results offer a 20-year M&R plan, in which road condition remains in good and fair condition range. The proposed approach can be used by decision-makers and transportation agencies to plan efficient maintenance and rehabilitation actions that can save costs and minimize environmental impacts.

# 1. Introduction:

The performance of all national economies is dependent on transportation infrastructure through a wide range of economic and social advantages. Additionally, the construction, maintenance, and use of transportation networks generally impose significant environmental impacts (Santero and Horvath, 2009). The need for sustainable infrastructure systems, such as transportation networks, is now widely acknowledged. In one of the most recent and most comprehensive studies, the UN's International panel on climate change (IPCC) reported for "code red" in its latest report (Cartwright, 2021). Accordingly, due to the high carbon footprint of construction industry, maintaining and repairing different components of the continually deteriorating transportation networks (e.g., pavements) with minimal greenhouse gas emissions, financial expenditure, and non-renewable materials consumption is now one of the most important priorities of infrastructure management authorities (Wang and Gangaram, 2014).

In this context, Pellicer et al. (2016) found that conventional methods of planning pavement maintenance alternatives concentrated on economic and technical parameters while ignored environmental impacts. Air pollution and climate change are two of the most pressing environmental issues facing the globe today. If temperature levels continue to rise owing to anthropogenic greenhouse gas (GHG) emissions into the atmosphere, global warming might cause severe environmental, social, and economic consequences across the world (Santos et al., 2020). Therefore, special attention must be paid to the transportation sector since it is responsible for spewing considerable amounts of pollutants and GHG up into the air (Watts et al., 2018). Moreover, in the presence of global warming, more M&R practices may be needed because of the increasing frequency and intensity of natural disasters (Khandel and Soliman, 2019). By conducting more number of M&R actions, even more pollutants will be released into the atmosphere (Santos et al., 2020). Accordingly, there is a need for comprehensive and efficient methods that can assist in planning optimal M&R plans that not only consider financial constraints but also cover environmental aspects.

Developing a practical long-term maintenance framework will not only help policymakers to achieve an acceptable level of service but enables them to avoid significant capital waste since there are budget limitations (France-Mensah and O'Brien, 2018). In recent years, many budget allocation models have been proposed to address this issue for various purposes on project and network levels (e.g., Yu et al., 2015). In this context, some of the studies in the field of pavement performance evaluation used local data sets (e.g., Yao et al., 2019) ans some other used the Long-Term Pavement Performance (LTPP) data (e.g., Piryonesi and El-Diraby 2020; Yamany et al., 2020).

To determine which maintenance technique results in best result, many researchers used the notion of "cost-effectiveness" to evaluate the efficiency of their methods. For instance, Torres-Machi et al. (2017) and Osorio et al. (2018) employed cost-effectiveness or cost-benefit techniques to consider performance and economy aspects in maintenance planning. Review of the available research (e.g., Alam et al., 2020; Hua et al., 2021; Irfan et al., 2012) in the area of pavement repair and maintenance also shows that combining treatments and developing more predictable schedules would considerably increase pavement service life and save a significant amount of capital. Additionally, to make better decisions that are cost-effective and more environmentally friendly, pavement repair must be planned over a lengthy period by considering the long-term pavement performance deterioration (Donev and Hoffmann, 2020; Li et al., 2021; Qiao et al., 2021; Renard et al., 2021).

Numerous researchers have attempted to utilize machine learning (ML) to tackle various engineering challenges, motivated by machine learning's incredible capacity to employ models built on data in order to replicate human intelligence activities (Neves et al., 2017; Chopra et al., 2018; Deka, 2019; Sheikh et al., 2021; Khandel et al., 2021a,b; Naranjo-Pérez et al., 2020). Among the available ML tools, deep learning is one the most prevalent which has been used in several applications including damage recognition, crack detection, and performance prediction in buildings (e.g., Cheng et al., 2021; Attari et al., 2017), bridges (e.g., Assaad and El-adaway, 2020a; Asghari and Hsu, 2022), roads (e.g., Donev and Hoffmann, 2020; Piryonesi and El-Diraby, 2020) sewer and power lines (e.g., Sabour et al., 2021; Wang et al., 2020), and dams (Assaad and El-adaway, 2020b).

Among different machine learning techniques, reinforcement learning (RL) has gained a lot of attention in recent years. RL can be used to provide efficient maintenance plans that can assist in long-term decision-making (Yao et al., 2020). Pavement authorities generally focus on the long-term cost-effectiveness of maintenance treatments which makes maintenance planning a sequential decision-making problem. However, by employing either professional expertise or mathematical programming approaches, long-term maintenance planning is still challenging. This is mainly due to the complex process of pavement deterioration, multiple alternative treatments, and various pavement performance indicators. RL is an excellent tool for simulating sequential decision-making situations on a big scale. It discovers the best approach by maximizing the agent's cumulative rewards from the environment. RL-based pavement management decision-making techniques can include sophisticated pavement performance prediction, more specific treatment techniques, and wide selection of materials in a long-term optimization through agent-action environment interactions, as opposed to traditional pavement management decision-making techniques (Liu et al., 2020, Han et al., 2021, Renard et al., 2021). Recently, some research has focused on finding the best treatment timing through maximizing cost-effectiveness using RL approaches.

Pavement designs and maintenance have historically used life cycle cost analysis (LCCA). The LCCA is a technique for calculating total cost based on various initial construction costs, maintenance costs, and vehicle fuel costs to achieve maximum net savings over the life cycle. This approach does not quantify the environmental impact across various stages of the life, but rather estimates the financial impact over the life of an investment. In contrast, life cycle assessment approach (LCA) is a tool for assessing the environmental implications of all phases of a product or process, from cradle to grave. LCA model is used to estimate the environmental damage cost (EDC) and construct life cycle inventories (LCIs). The inventory of material, energy, and environmental pollutant flows from and to nature are some of the most common LCIs. Additionally, different modules including material, transportation/distribution, construction, congestion due to M&R activities, usage, and end of life (EOL) modules are typical components of a comprehensive LCA model.

In this context, Labi and Sinha (2005) investigated the economic effectiveness of various degrees of life cycle preventive maintenance for three asphaltic concrete pavement functional class families and developed a generic framework for evaluating investment alternatives. Mandapaka et al. (2012) created an integrated LCCA and California mechanistic-empirical design procedure model to analyze and select an effective M&R plan. Praticò et al. (2011) created an LCCA model based on a conceptual engineering economic tool to handle agency, user, and externality costs for solutions to stabilize unstable subgrade soils for low-volume highways. The previously created LCCA models take several pathways to account for the economic features of pavements, with an emphasis on the agency and user costs. Several other researchers focsed on quantifying environmental impacts using LCA approach. For instance, Keoleian et al. (2005) calculated the environmental implications of two bridge deck systems considering material manufacturing, transportation, building and maintenance procedures, congestion impact, and EOL treatment. Huang et al. (2009) developed a system for constructing and maintaining asphalt pavements using an LCA model that is fully characterized in terms of material and construction modules. Santero and Horvath (2009) devised a life cycle assessment technique to assess the effect of global warming on the key components of pavements over their life cycle. Wang et al. (2012) assessed the energy consumption and greenhouse gas (GHG) emissions of several pavement rehabilitation options for material manufacture, construction, and usage, with a specific emphasis on the impacts of pavement rolling resistance on vehicle fuel consumption.

To benefit from the advantages of both LCCA and LCA approaches in pavement M&R planning, the environmental aspects can be addressed by LCA while the LCCA approach can cover the economic aspects. An integrated LCA–LCCA model can be significantly beneficial in developing an efficient cost evaluation system with improved pavement design and maintenance techniques. Accordingly, in this paper, an RL-based framework which integrates LCA and LCCA is developed to determine the optimum type and timing of M&R practices based on the LTPP data. The aim of the proposed framework is to find a cost and environmentally-effective sequence of maintenance actions to provide the most budget-efficient set of maintenance actions and contribute to a more sustainable maintenance management system. The significant contribution of this paper may be summarized as follows: (a) for the first time, this paper is developing an RL framework that considers costs and environmental effects at the same time, and (b) developing an RL framework based on a public database (LTPP) is the first one in its kind.

## 2. Deep Reinforcement Learning

Reinforcement learning is a branch of machine learning specialized in solving sequential decision-making problems to allow software-defined agents (algorithms) to learn the optimum behaviors in a virtual environment to achieve their objectives (Barto, 2014). The main difference of RL compared to other machine learning tools is the type of feedback (reward) that the agent receives after making a decision. In supervised learning, the agent simply understands how good it was in predicting (doing an action) by immediately comparing its result with the label (Cunningham et al., 2008). The agent then uses this comparison to improve its next prediction (action). For unsupervised learning, there is no feedback as the data is unlabeled (Ghahramani, 2003). Reinforcement learning uses an alternative method which works with *delayed rewards*. In this method, agents are learning through a trial-and-error procedure of interacting with the environment. The method starts with an initial state, and every time the agent makes a decision, it receives a reward and updates the state. The goal of the agent is to find a *Policy* that leads to the maximum reward. RL works on the Markov Decision Process (MDP) principles which can be shown as a tuple $M = (S, A, T, d_0, r, \gamma)$, where S is a set of states s ∈ S, A is a set of actions that can be discrete or continous $a \in A$, $T$ is conditional probability distribution of the form $T(s_{t+1} | s_t, a_t)$ that describes the dynamics of the system, $d_0$ is the initial state distribution $d_0(s_0)$, $r: S \times A \to R$ is the reward function, and $\gamma \in (0,1]$ is thescalar discount factor. Figure 1 shows the schematic view of a RL system. In a model-free prediction, $T$ is always unknown and is thus omitted afterwards in this paper.

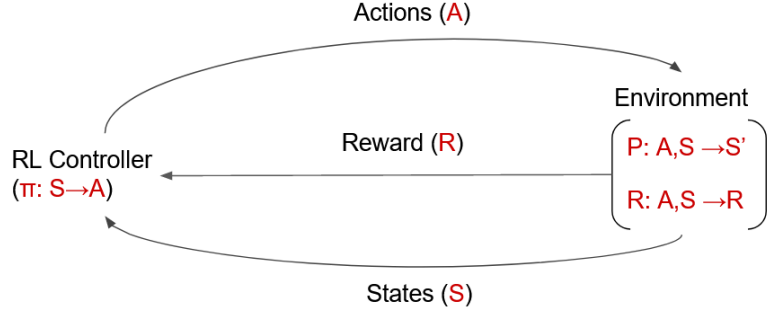

*Figure 1. the schema of the flow in an RL system*

The best (optimal) policy for RL is one that maximizes the cumulative reward. At each step in policy π the agent at one specific state $s_t$ takes an action $a_t$ to observe the next resulting state $s_{t+1}$ and obtains a reward $r_{t+1}$. The final cumulative reward in step $t$ is defined by $G_t$ and can be defined as follow:

$$G_t = \gamma^0 r_{t+1} + \gamma^1 r_{t+1} + ... + \gamma^{n-1} r_n$$

where $n$ is the number of steps. As mentioned above, $\gamma$ is discount factor, and it plays a key role in "delayed reward" phenomenon. The more $\gamma$ is closed to 0, the less important are future rewards, and the agent focuses on immediate rewards (Sutton 1992). Achieving an optimal policy $\pi_*$ means that the agent has succeeded in finding the optimal sequence of actions in each time step which maximizes $G_t$.

## 3. Proximal Policy Optimization (PPO)

PPO is an *on-policy* method that uses policy gradient approaches to estimate the optimum policy. It makes the use of *importance sampling* and tends to increase the probability of sampled actions through a gradient ascent approach. PPO make adaptive updates with its *clipped surrogate objective*. It also ensures that the deviation from the previous policy is relatively small. The objective is as follows:

$$Loss^{CLIP}(\theta) = E_t[\min(\frac{\pi_\theta(a_t | s_t)}{\pi_{\theta_{old}}(a_t | s_t)} Adv_t \ , \ \text{clip}(\frac{\pi_\theta(a_t | s_t)}{\pi_{\theta_{old}}(a_t | s_t)}, 1-\alpha, 1+\alpha) Adv_t]$$

Where $\pi_\theta$ is the updated policy that derives from the old policy $\pi_{\theta_{old}}$ and $\alpha$ is a hyperparameter that ranges between 0 and 1. The clipping removes the incentive for moving $\frac{\pi_\theta(a_t|s_t)}{\pi_{\theta_{old}}(a_t|s_t)}$ outside of the interval $[1-\alpha, 1+\alpha]$ and penalizes the objective for having too large of a policy update. Adv is the Advantage function which represents a measure of acceptability, in a certain state.

The pseudo-code of PPO is as follows:

---
**Algorithm 2** Proximal Policy Optimization
---
for episode = 1, M do
    for executer 1, N do
        run policy $\pi_{\theta_{old}}$ in environment for $t$ time steps and calculate $Adv_t$ in each time step
    end for
    calculate $Loss^{CLIP}(\theta)$ with K epochs on a samples mini batch and update $\theta$
end for

---

## 4. Environment

The Environment is the world that the agent moves and interacts with. The Environment receives the agent's current state and action as inputs and delivers the reward and the agent's future state as output.

In this sustainable maintenance planning system, the Environment acts as a place that simulates the initial road condition and then determines the following condition of the road based on the chosen action. In a perfect world, the agent can perform its action on an actual road segment and quickly evaluate the post-action condition of the road to evaluate the effectiveness of the action. However, due to the complexity and required effort, employing such method in real-world applications and even their laboratory simulations may not be practical. To overcome this problem, Machine Learning models are used to simulate the Environment and estimate future road conditions. In this paper, a Deep Artificial Neural Network (DANN) is employed to estimate the following condition of a road section based on some initial (input) features. The DANN model is then used to play the role of the Environment for RL. Figure 2 depicts the structure of the employed DANN. As shown, the DANN model comprise of one input layer, three blocks and an output layer. Each block is a sequence of a dense layer (fully connected layer), a Parametric Rectified Linear Unit (PRelu) activation function, and a Dropout layer. The PRelu layer is responsible for activating or deactivating Neurons while a Dropout layer randomly deletes the connection of some neurons (Dropout 0.3 deletes the connection for 30 percent of neurons).

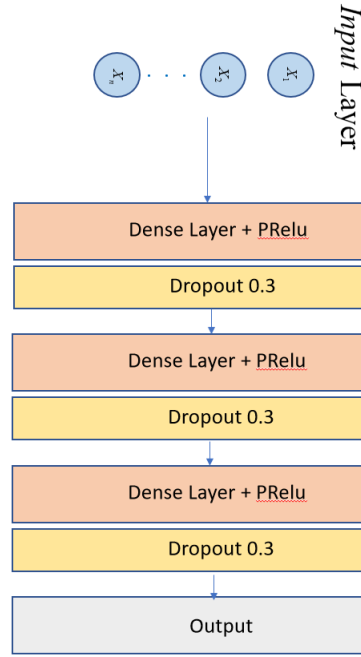

*Figure 2. the Flowchart of the DNN model that predicts the indicator performance and is used as the Environment in the RL*

## 5. Data

Factors related to road structure features, traffic conditions, weather conditions, initial performance, interval between two evaluations, maintenance types and time are used as the input of for the DANNs. Two indicators are considered in the Sustainable Maintenance Planning Model, two individual DANNs are used to separately predict each indicator. Long-term pavement performance (LTPP) data are adopted in this paper. LTTP contains several indicators related to flexible pavements (e.g., International Roughness Index (IRI), Rutting Depth (RD), Cracking Metric (CM), and Faulting (F)). LTPP database is recognized as one of the most comprehensive dataset of road assets. However, this dataset still has deficiencies such as missing and/or conflicting data. To deal with these shortcomings, several approaches have been employed to improve the quality of the data set. Additionally, several studies proposed methods to develop secondary indicators based on the indicators that are already included in the LTPP database (e.g., Piryonesi and El-Diraby, 2020). In addition, Based on the research conducted by (Lupia and Alter 2014), the transparency of the data is important; thus, it is worth mentioning that the research in this paper is conducted based on an open-source dataset (as opposed to previous similar work), that may enhance the validity of the proposed methods.

## 6. Indicator Selection

Choosing the right indicator has a significant impact on the outcome and may turn a credible, extensive, and complete maintenance system into a meaningless one. Furthermore, selecting a single indicator can result in inaccuracies since it may not offer a comprehensive picture of road condition. As a result, a collection of indicators can assist in accurately reflecting the road condition. Based on the review of different selection criteria proposed in the literature (e.g., Domitrović et al., 2018; Marcelino et al., 2018; Visintine et al., 2018; Fakhri and Dezfoulian, 2019;

Nabipour et al., 2019; Li et al. 2020; Cruz et al., 2021; Fakhri et al., 2021; Song et al., 2021) a reliable performance indicators should be comprehensive, balanced, capable of showing trends, economical to measure, and appropriate. Table 2 presents the definition of each of those criteria.

Table 1– Indicator selection Criteria

| Criteria | Definition |
|---|---|
| **Comprehensive** | Includes all key aspects of performance (addresses both functional and structural performance) |
| **Balanced** | Includes several types of metrics |
| **Able to show trends** | Can be used to analyze the performance over time |
| **Economical to measure** | Does not increase burden on a State to collect/measure |
| **Appropriateness** | Is suitable as a measure at the national level (including accuracy, timeliness, consistency, and precision) |

Among the available indicators, RD is main driver of the overall road condition in wet and non-freeze climate conditions (Visintine, Simpson et al. 2017). Moreover, IRI has continuously been chosen as the main indicator of road and highway condition by different studies. Therefore, IRI and RD are selected as the two indicators in this paper. Table 3 presents condition thresholds for these two indicators.

Table 3– Pavement condition rating thresholds

| Condition Metric | Performance Level | Thresholds |
|---|---|---|
| IRI | Good | <95 |
| IRI | Fair | 95–170 |
| IRI | Poor | >170 |
| RD | Good | <0.20 |
| RD | Fair | 0.20–0.40 |
| **RD** | **Poor** | **>0.40** |

# 7. States

A state is a situation that the agent evaluates itself in the environment. The initial state updates itself to a new state by performing a series of actions. In this paper, 19 features combined (representing structural, traffic, and climate conditions) with the two performance indicators (I.e., IRI and RD) form the state space. Table  shows the features and their categories. This set of features reflect the road attributes that will change during the maintenance horizon.

Table 4. Input features (states) of the DNN model of RL algorithm

| Category of Feature | Features (states) |
|---|---|
| **Structure** | Original surface type |
|  | Original surface thickness |
|  | Original surface material |
|  | Binder type |
|  | Binder thickness |
|  | Binder material |
|  | Base type |
|  | Base thickness |

|  | | |
|---|---|---|
| | Base material | |
| | Subbase type | |
| | Subbase thickness | |
| | Subbase material | |
| **Traffic** | Truck Ratio | |
| | Annual EASL | |
| | Annual AADT | |
| **Climate** | Annual precipitation | |
| | Freeze thaw | |
| | Freeze/Non-freeze | |
| | Wet/Dry | |

# 8. Actions

All motions that the agent can take are known as Actions (A). Although an Action is relatively self-explanatory, agents typically pick from a list of discrete and viable acts.

Table shows Action type, thickness and material types that the agent considers. These action types represent more than 70% of the action types practiced in the LTPP data set. "Do-nothing" is also considered to be an action type. Different combinations of types, thickness, and materials are then used to generate individual Actions that form the Action space. This helps to reach highly accurate results with the minimal computational effort.

Table 5. Possible Actions for the agent

| Type of Action | Thickness | Material |
|---|---|---|
| **Asphalt concrete overlay** | 25.4 | AC-20 |
| | 50.8 | |
| | 76.2 | AC-30 |
| | 101.6 | |
| **Hot-Mix Recycled Asphalt Concrete Overlay** | 38.1 | AC-20 |
| | 50.8 | AC-30 |
| | 76.2 | |
| **Mill Off AC and Overlay with AC** | 38.1 | AC-20 |
| | 50.8 | |
| | 76.2 | AC-30 |
| | 101.6 | |
| **Mill Existing Pavement and Overlay with hot-Mix Recycled AC** | 50.8 | AC-20 |
| | 76.2 | AC-30 |
| | 101.6 | |
| **Aggregate Seal Coat** | N/A | Asphalt Cements AC-20 |
| **Fog Seal Coat** | N/A | HFRS-2P (POLYMERIZED) |
| **Crack sealing and Patching** | N/A | AC-20 |
| Do-nothing | N/A | N/A |

# 9. Reward

A reward is feedback that is utilized to determine if an agent's activities in a particular state were successful or unsuccessful. The purpose of incentives is to offer feedback on an RL model's actions' performance. As a result, it is critical to define the reward accurately to direct the learning process properly. The reward in this paper is proposed based on the idea in a previous research (Yao, Dong et al. 2020). However, the reward function differs from the previous work in several terms, including the environmental cost. Figure 3 represents the calculation of actions' performance on an indicator, in which the initial point is the first year of planning. The purple curve shows the indicator performance after taking an action. The blue dotted area is the area that is used in the reward calculation formula, and the black curve is a hypothetical curve of not doing an action at all during the planning horizon. When the planning starts, the indicator is in an initial state. However, each year it decreases if no maintenance action is performed on the road, and it follows the black curve. The first time an action is performed (purple curve), it increases and separates from the non-action curve. Generally, the performance will decrease between two maintenance actions (time $t$ and $t+1$). It is worth noting that for the sake of simplicity, Figure 3 shows a decreasing performance indicator, while the proposed ones (IRI and RD) are inherently increasing.

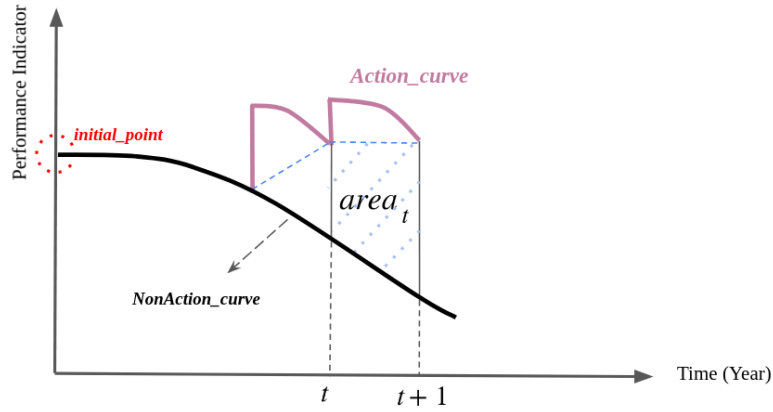

*Figure 3. The mechanism of calculating the effectiveness of actions for an indicator*

The proposed reward function is as follows: In each time step, firstly, the cumulative area bounded with non-action curve, Action curve, current time $t$ and the next time $t+1$ is calculated for each indicator (eq 1). In eq1, $i$ indicates the indicator.

$$total\_area_t^i = \sum_{t=t_0}^{t} area_t^i \quad eq\ 1$$

In addition, the total cumulative cost is calculated as in eq 2. In this paper, as the maintenance system is considered to be both environmentally and economically effective, the total cost is the sum of action cost and environmental cost. In eq2, 1.04 is for 4% discount rate as it has been largely used in the previous research. Furthermore, the environmental cost of each action is calculated based on eq 3, in which $\vartheta_{action_t}$ and $\varsigma$ show the equal amount of CO2 (metric ton) (Table and Figure 7) and marginal cost of carbon dioxide (Tol 2005) respectively.

$$total\_cost_t = \sum_{t=t_0}^{t} (\frac{cost_{action_t} + env\_cost_{action_t}}{1.04^t}) \quad eq\ 2$$

$$\text{env\_cost}_{\text{action}_t} = \vartheta_{\text{action}_t} * \varsigma \quad eq\ 3$$

In the next step, the effective cost is calculated based on eq 4, which is the effectiveness of applying an action (including non-action) so far in the system.

$$\text{eff\_cost}_t^i = \frac{\text{total\_area}_t^i}{\text{total\_cost}_t} \quad \text{if}\ \text{total\_cost}_t \neq 0\ \text{else}\ 0 \quad eq\ 4$$

The effective cost for each indicator is then combined into a single term, as shown in eq 5.

$$\text{final\_effcost}_t = \omega_{IRI} \times \text{eff\_cost}_t^{IRI} + \omega_{RD} \times \text{eff\_cost}_t^{RD} \quad eq\ 5$$

$\omega_{IRI}$ and $\omega_{RD}$ are considered 55 and 45 percent, respectively. Finally, the reward is defined in eq 6.

$$\text{reward}_{t+1} = \text{final\_effcost}_{t+1} - \text{final\_effcost}_t \quad eq\ 6$$

## 10. Case Study and Results

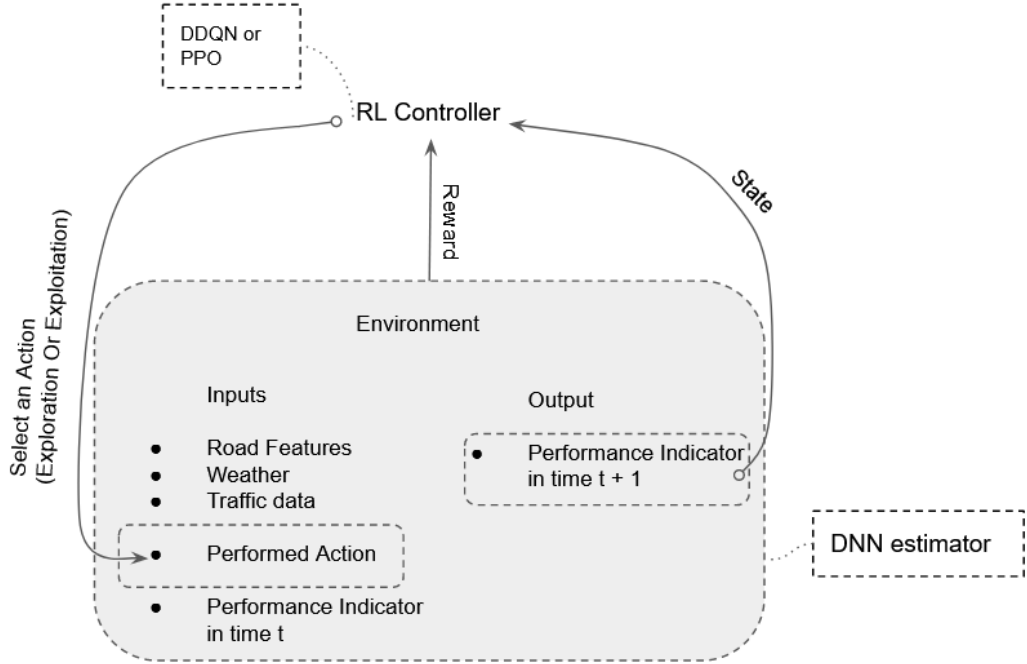

*Figure 4. the schema of the proposed sustainable maintenance planning system*

In this section the overall Flowchart of the proposed research has been explained. As illustrated in Figure 4, the RL controller passes the selected action based on the state and reward which were calculated in the Environment. Consequently, the selected action is used again in the Environment to calculate the Reward and next state. As mentioned in the previous section, a DNN is used as the main part of the Environment. With this DNN, the Reward and next states can be calculated. The architecture of the proposed DNN was explained in the previous section. However, in order to prepare the data to be used in this architecture, some data analysis techniques must be employed on the data.

**Data Preparation.**

There are 40 factors used as the inputs for the ANN model, which are related to road structure features (24 factors), traffic conditions (6 factors), weather conditions (4 factors), initial performance indicator (1 factor), the interval between two indicator evaluations (1 factor), maintenance type (3 factors) and maintenance date (1 factor).

A major and time-consuming task for academic research papers dealing with big databases is the data preparation part (García, Luengo et al. 2015). As the required dataset for training the next-condition predictor model described in the above section is based on some separate data sources (LTPP database and climate module MERRA provided by NASA), these data need to be first merged into a single one. Data preprocessing is also a must-do for the neural network prediction model (Environment). Without preprocessing of raw data, the neural network will not produce accurate forecasts.

The main tasks of the preprocessing part are as follows: First, the categorical data (e.g., road function class, Surface type, etc.) are converted into numerical ones, using the one-hot encoding method. One-hot encoding is a way to assign a numerical representation to categorical data in order to make them feasible to participate in mathematical models. Second, a Min-Max Normalizer is applied to the data for faster convergence and higher proficiency (Patro and Sahu 2015). Furthermore, considering traffic and weather data, there are some missing data, which can be handled by interpolation, *fillna* methods and deletion. In addition, because of some illogical cases in the dataset, IRI and RD might have decreased over time without any actions taken. To overcome this problem, a calibration method has been applied on those data points that have decreased between two action-performed dates (if two maintenance actions have been done on a single road segment, its indicator performance must increase gradually in this interval. Otherwise, it is erroneous and must be calibrated). The calibration process is illustrated in Figure 5, in which The blue squares are actual data. The red vertical lines show the time when a maintenance action has been done. In a rational situation, the performance indicator like RD decreases right after a maintenance action but then increases. Thus, those data points that decrease without performing maintenance action are wrong and must be calibrated. In this figure, the calibrated data points are indicated with green cross-shapes.

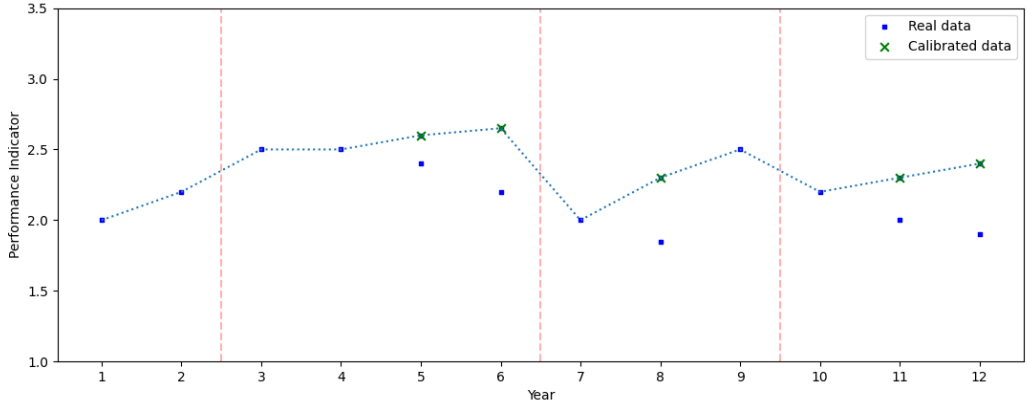

*Figure 5. The calibration process of the data.*

**DNN Results.** After Preparing the data, it is then fed into two DNN models to produce the outputs. The outputs (IRI and RD) are the predicted performance indicators which are calculated based on the current features of the road and a maintenance action. The performance of this DNNs are shown in Figure 6, which represents the real data with red and the predicted data with blue lines for 50 random samples.

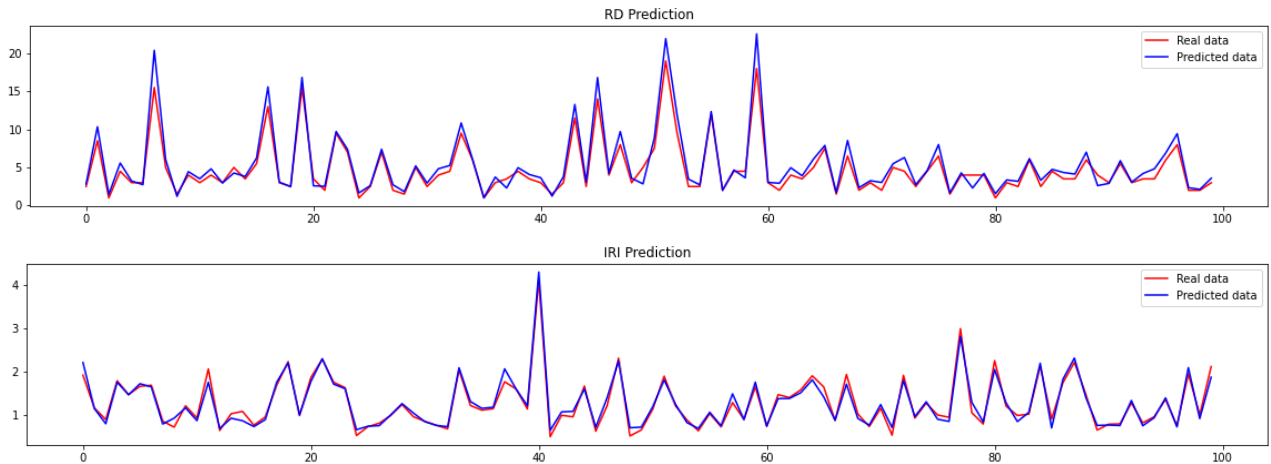

*Figure 6. The performance of the prediction model in predicting the performance indicators*

1. **PPO Hyper Parameters**

It can be observed that PPO provides a better convergence and performance rate than other techniques but is sensitive to changes. On the other hand, DQN alone is unstable and gives poor convergence, hence requiring several add-ons.

The obtained hyperparameters are shown in Table .

Table 6– Model Hyperparameters for the PPO algorithm

| Parameters | Values |
|---|---|
| Planning horizon | 20 years |
| Beginning year | 2021 |
| Number of steps M | 128 |
| Number of Executors N | 4 |
| Learning rate | *0.00025* |
| Discount factor | 0.99 |
| Entropy coefficient for the loss calculation | 0.04 |
| Value function coefficient for the loss calculation | 0.5 |
| Clip range $\alpha$ | 0.2 |

## 2. Integrating LCA and LCCA

To calculate the cost-effectiveness of the actions, first, the cost and environmental impact of treatments should be calculated. The environmental damage cost (EDC) has generally been overlooked when calculating pavement costs. To consider EDC alongside economic cost, there should be an integration between LCA and LCCA. A model provided by Tol (Tol 2005) has been utilized for the integration. The model takes the price of capturing a ton of CO2 as the environmental cost. However, there are wealth of literature that calculates this cost and it varies between 5 $/tC and 1667 $/tC. The authors of this paper are not experts in this field and considered the mean value in the investigated literature as a basis for the study. After this step, the cost and environmental impact (as marginal costs) of the M&R actions are integrated. In the following paragraphs, a discussion about the implemented approaches is presented.

## 3. LCA and LCCA

For the LCA, a tool developed by a consortium on green design and manufacturing in the University of California, Berkeley (paLATE 2.0) (Figure 7) is used with some modifications. The tool takes user input for the design, initial construction, maintenance, equipment use, and costs for a roadway, and provides outputs for the life-cycle environmental effects and costs. In this research, only production of material, transportation and construction have been considered. Other parts are simply beyond the scope of this research. Environmental effects investigated include CO2, CH4, N2O, PM2.5, SO2, CO and NOx. All of these pollutants have been considered in LCA as an equal amount of CO2, based on the global warming potential (GWP). In Table , Factors for different pollutants are provided (Jing, Bai et al. 2012). In the following parts, more details will be discussed:

Table 7– Equal amount of CO2

| Pollutants | GWP (CO2-eq) | AP (SO2-eq) | REP (PM2.5-eq) |
|---|---|---|---|
| SO2 | - | 1 | 1.9 |
| CO2 | 1 | - | - |
| NOx | - | 0.7 | 0.3 |
| PM2.5 | - | - | 1 |
| CO | 3 | - | - |
| CH4 | 21 | - | - |
| N2O | 310 | 0.7 | - |

To calculate economic and environmental impact of M&R practices, paLATE tool is used. As it's demonstrated in figure 7, the user can input the details about different layers of pavement and the tool provides the cost and produced pollution during three phases of production, transportation, and construction. The results then, compared to other research and were in a reasonable range.

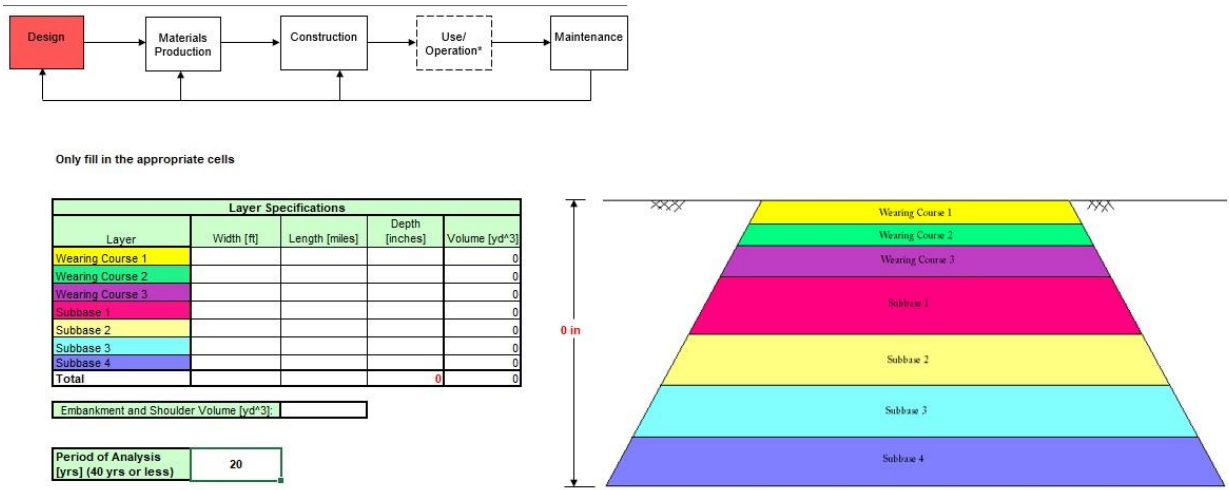

*Figure 7– paLATE 2.0 tool*

**Production**. Pavement material manufacturing emits a lot of pollutants into the atmosphere. For the estimation of the environmental impact of the material production process, research by European Bitumen Association is used (Euro bitumen 2012).

**Transportation.** The transportation module tracks the movement of paving materials from the plant to the construction site, as well as the recycled pavement materials to the recycling plant as they reach the end of their useful lives. The air emissions for material transportation were modeled using the (GREET, 2010) model.

**Construction.** Construction equipment emits pollutants during maintenance and rehabilitation. The model provided by The NONROAD model (U.S. EPA, 2008) has been used to estimate the emissions based on hourly consumption of equipment.

It is important to note that the social impacts of the road maintenance and rehabilitation were not included in this paper because they were out of the scope of this study.

### 4. Case study

The case study is a hypothetical 6 lane (3 lanes in each direction) highway located in Texas. The highway is set to be in a warm and wet climate. Daily traffic conditions (AADT, ESAL, different axle counts, Truck ration) are compatible with the road structure. The highway is composed of 46 sections and a total length of 23 kilometers for ease of calculation, all road segments are considered to be 500 meters in length (the length of the road segments are similar to the case study in Yao et al. study (Yao, Dong et al. 2020)), and the width of all lanes is 3.66 meters (46 segments in total). The reason for selecting a hypothetical case study is to implement a comprehensive set of characteristics in the case study to compare the performance of the model under different conditions. Different conditions for the case study are defined within the limits of similar highways in the same region. Road sections were classified as "wet" because the average site precipitation was greater than 20 inches/yr. Moreover, they are classified as "freeze" since the average site freeze index was greater than 46 °C – degree days.

Figure 8 denotes the learning process for the agent, which takes thousands of iterations. Each iteration or round is a set of 20 years of performing action on a road segment. As it can be seen the model converges after iteration 7000.

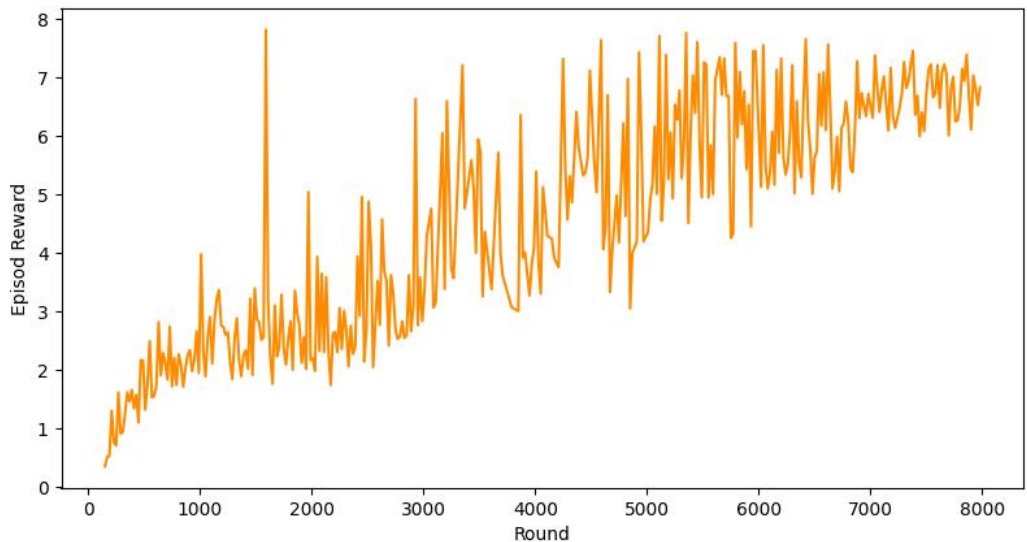

*Figure 8. Episode reward of each round during PPO training*

Figure 9 shows the projected maintenance and rehabilitation costs of all segments between 2022 and 2041. It is important to note that more budget is needed in some years. This is because the road is in an excellent overall condition in the early years, and thus, there is no need for heavy M&R practices. After several years that the road has been subjected to heavy traffics, the need for heavy M&R practices increased. Since after the heavy M&R, the condition of the road improves, there is a sharp decline in budget allocation. In Figure 10, which is only for a single road segment, it is obvious that in some years, there is no M&R, which happens after heavy rehabilitation practices. This means that the model can decide that there is no need for more M&R practices after recent rehabilitation.

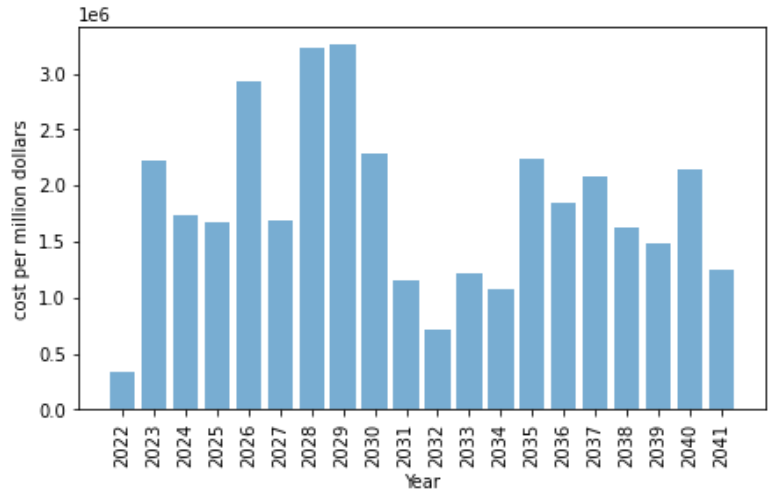

Figure 9 Total M&R costs

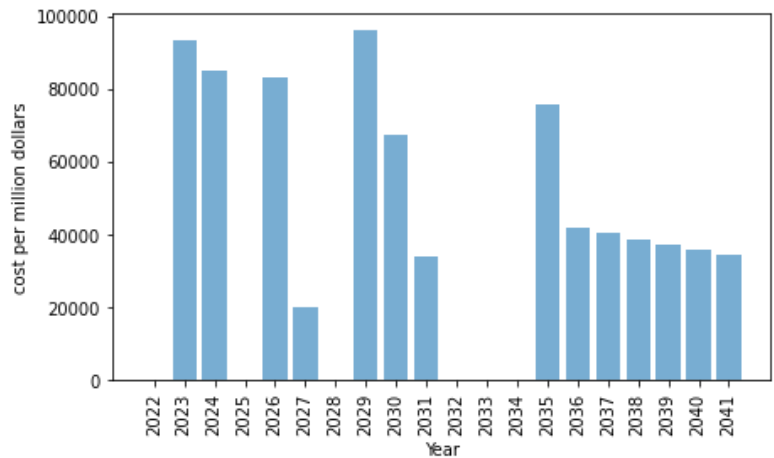

Figure 10– M&R costs for one segment

Table shows the percent of actions for each year. Since the road is in good condition at the start of 2022 (the beginning of the 20-year period), 60 percent of the first year's actions are non-actions. Moreover, the other 40 percent is a slight maintenance practice (aggregate seal coat). After the first year, the road will be subjected to heavy traffic, and the need for heavy M&R practices is necessary.

In order to see if the model responds logically to different conditions, a sensitivity analysis has been conducted. Figure 11 shows that as the traffic volume increases (by 50, 100, 150, 200, and 250 percent), the need for maintenance and rehabilitation increases, which results in a drop in the cost-effectiveness of the model. It can be concluded that the developed model responds logically.

*Table 8– Percent of road M&R practices per year*

| Time (year) | Asphalt Concrete Overlay | Hot-Mix Recycled Asphalt Concrete Overlay | Mill Off AC & Overlay w/ AC | Mill Existing Pavement & Overlay w/ Hot-mix Recycled AC | Aggregate Seal Coat | Fog Seal Coat | No Action |
|---|---|---|---|---|---|---|---|
| 2022 | 0 | 0 | 0 | 0 | 40 | 0 | 60 |
| 2023 | 85 | 0 | 0 | 0 | 15 | 0 | 0 |
| 2024 | 0 | 27 | 6 | 0 | 21 | 0 | 46 |
| 2025 | 35 | 25 | 0 | 0 | 21 | 0 | 19 |
| 2026 | 29 | 54 | 15 | 2 | 0 | 0 | 0 |
| 2027 | 65 | 23 | 6 | 0 | 0 | 4 | 2 |
| 2028 | 0 | 8 | 46 | 0 | 6 | 0 | 40 |
| 2029 | 0 | 27 | 52 | 0 | 6 | 2 | 12 |
| 2030 | 8 | 48 | 8 | 0 | 19 | 2 | 15 |
| 2031 | 31 | 4 | 2 | 2 | 40 | 0 | 21 |
| 2032 | 21 | 0 | 4 | 0 | 38 | 0 | 38 |
| 2033 | 0 | 4 | 19 | 0 | 48 | 0 | 29 |
| 2034 | 0 | 6 | 12 | 0 | 50 | 0 | 31 |
| 2035 | 31 | 15 | 29 | 0 | 10 | 15 | 0 |
| 2036 | 31 | 10 | 29 | 0 | 6 | 23 | 0 |
| 2037 | 56 | 0 | 27 | 0 | 17 | 0 | 0 |
| 2038 | 83 | 0 | 0 | 0 | 17 | 0 | 0 |
| 2039 | 77 | 0 | 0 | 0 | 23 | 0 | 0 |
| 2040 | 77 | 0 | 0 | 17 | 6 | 0 | 0 |
| 2041 | 67 | 0 | 0 | 0 | 33 | 0 | 0 |

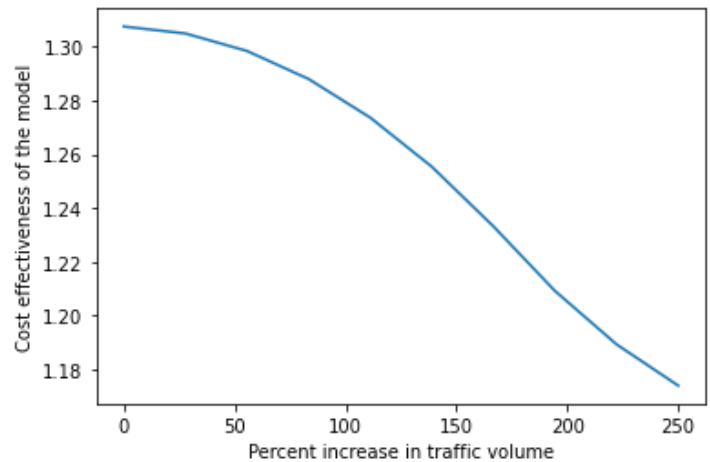

*Figure 11. Sensitivity Analysis of cost-effectiveness due to increase in Traffic volume*

## 11. CONCLUSION

This study introduced a new framework for long-term rehabilitation and road maintenance strategies. A PPO model has been utilized to maximize the cost-effectiveness of actions. This tool can provide a basis to help decision-makers. The result shows that the proposed framework can successfully produce promising results. A DQN model is also tested, but PPO results were more accurate and logical. After each intervention, all lanes must reach the same elevation. Since the real-time practice for measuring the impact of traffic and M&R practices is not achievable, an ANN model to simulate the real-world condition is used. The environment comprises 40 parameters divided into four subcategories of 1) Structure, 2) Traffic, 3) PIs and, 4) Climate. The action space is made up of 32 actions (including do-nothing) which only 16 of them are taken by the agent. The main features of the study are as follows: 1) The proposed framework solves the problem of conflicting data by correcting the false data. 2) A neural network model specifically created to simulate the environment. 3) A sensitivity analysis was conducted to evaluate the proposed model's behavior. 4) The model integrates the environmental and economic impact of M&R practices to produce a more comprehensive basis for decision-makers. 5) the outcome of the research is a decision support tool for organizations to go towards M&R strategies that align with sustainable development goals.

Future studies could address the impact of climate change and evaluate how different climate change scenarios can affect the M&R strategies. Another direction can be pushing models toward a network of roads instead of considering one segment of road. Incorporating budget restrictions into the model will lead to more realistic results since there are always funding restrictions on the side of organizations responsible for M&R practices in the real condition. Finally, implementing the performance indicators in the reward calculation, so that different service levels have different values in the reward will lead to a more logical behavior of the model.

## References

Alam, M. R., et al. (2020). "A systematic approach to estimate global warming potential from pavement vehicle interaction using Canadian Long-Term Pavement Performance data." Journal of Cleaner Production **273**: 123106.

Andrew Barto, R. S. S. (2014). <u>Reinforcement Learning: An Introduction</u>.

Asghari, V. and S.-C. Hsu (2022). "Upscaling Complex Project-Level Infrastructure Intervention Planning to Network Assets." <u>Journal of Construction Engineering and Management</u> **148**(1): 04021188.

Aslani, M., et al. (2018). "Traffic signal optimization through discrete and continuous reinforcement learning with robustness analysis in downtown Tehran." <u>Advanced Engineering Informatics</u> **38**: 639-655.

Assaad, R. and I. H. El-adaway (2020). "Bridge infrastructure asset management system: Comparative computational machine learning approach for evaluating and predicting deck deterioration conditions." <u>Journal of Infrastructure Systems</u> **26**(3): 04020032.

Assaad, R. and I. H. El-adaway (2020). "Evaluation and prediction of the hazard potential level of dam infrastructures using computational artificial intelligence algorithms." <u>Journal of Management in Engineering</u> **36**(5): 04020051.

Athanasiou, A., et al. (2020). "A machine learning approach based on multifractal features for crack assessment of reinforced concrete shells." <u>Computer-Aided Civil and Infrastructure Engineering</u> **35**(6): 565-578.

Attari, N., et al. (2017). <u>Nazr-CNN: Fine-grained classification of UAV imagery for damage assessment</u>. 2017 IEEE International Conference on Data Science and Advanced Analytics (DSAA), IEEE.

Azimi, M. and G. Pekcan (2020). "Structural health monitoring using extremely compressed data through deep learning." <u>Computer-Aided Civil and Infrastructure Engineering</u> **35**(6): 597-614.

Cartwright, E. D. (2021). ""Code Red"—Recent IPCC Report Warns Time is Running Out on Climate Change." <u>Climate and Energy</u> **38**(3): 11-12.

Cheng, C. S., et al. (2021). "Deep learning for post-hurricane aerial damage assessment of buildings." <u>Computer-Aided Civil and Infrastructure Engineering</u>.

Chopra, P., et al. (2018). "Comparison of machine learning techniques for the prediction of compressive strength of concrete." Advances in Civil Engineering **2018**.

Cruz, O. G. D., et al. (2021). International Roughness Index as Road Performance Indicator: A Literature Review. IOP Conference Series: Earth and Environmental Science, IOP Publishing.

Cunningham, P., et al. (2008). Supervised learning. Machine learning techniques for multimedia, Springer**:** 21-49.

Deka, P. C. (2019). A primer on machine learning applications in civil engineering, CRC Press.

Domitrović, J., et al. (2018). "Application of an artificial neural network in pavement management system." Tehnički vjesnik **25**(Supplement 2): 466-473.

Donev, V. and M. Hoffmann (2020). "Optimisation of pavement maintenance and rehabilitation activities, timing and work zones for short survey sections and multiple distress types." International journal of pavement engineering **21**(5): 583-607.

Fakhri, M. and R. S. Dezfoulian (2019). "Pavement structural evaluation based on roughness and surface distress survey using neural network model." Construction and Building Materials **204**: 768-780.

Fakhri, M., et al. (2021). "Developing an approach for measuring the intensity of cracking based on geospatial analysis using GIS and automated data collection system." International journal of pavement engineering **22**(5): 582-596.

France-Mensah, J. and W. J. O'Brien (2018). "Budget allocation models for pavement maintenance and rehabilitation: Comparative case study." Journal of Management in Engineering **34**(2): 05018002.

García, S., et al. (2015). Data preprocessing in data mining, Springer.

Ghahramani, Z. (2003). Unsupervised learning. Summer School on Machine Learning, Springer.

Han, C., et al. (2021). "Asphalt pavement maintenance plans intelligent decision model based on reinforcement learning algorithm." Construction and Building Materials **299**: 124278.

Huang, M., et al. (2021). "LCA and LCCA based multi-objective optimization of pavement maintenance." Journal of Cleaner Production 283: 124583.

Irfan, M., et al. (2012). "Establishing optimal project-level strategies for pavement maintenance and rehabilitation–A framework and case study." Engineering Optimization 44(5): 565-589.

Jiang, S. and J. Zhang (2020). "Real-time crack assessment using deep neural networks with wall-climbing unmanned aerial system." Computer-Aided Civil and Infrastructure Engineering 35(6): 549-564.

Jiang, Z., et al. (2019). "Q-learning approach to coordinated optimization of passenger inflow control with train skip-stopping on a urban rail transit line." Computers & Industrial Engineering 127: 1131-1142.

Jing, Y.-Y., et al. (2012). "Life cycle assessment of a solar combined cooling heating and power system in different operation strategies." Applied Energy 92: 843-853.

Li, F., et al. (2021). Preventive Maintenance Technology for Asphalt Pavement, Springer.

Li, Y., et al. (2020). "A novel evaluation method for pavement distress based on impact of ride comfort." International journal of pavement engineering: 1-13.

Liu, Y., et al. (2020). "Dynamic selective maintenance optimization for multi-state systems over a finite horizon: A deep reinforcement learning approach." European Journal of Operational Research 283(1): 166-181.

Lupia, A. and G. Alter (2014). "Data access and research transparency in the quantitative tradition." PS: Political Science & Politics 47(1): 54-59.

Mannion, P., et al. (2016). "An experimental review of reinforcement learning algorithms for adaptive traffic signal control." Autonomic road transport support systems: 47-66.

Marcelino, P., et al. (2018). "Comprehensive performance indicators for road pavement condition assessment." Structure and Infrastructure Engineering 14(11): 1433-1445.

Mnih, V., et al. (2013). "Playing atari with deep reinforcement learning." arXiv preprint arXiv:1312.5602.

Müller, A., et al. (2021). "LemgoRL: An open-source Benchmark Tool to Train Reinforcement Learning Agents for Traffic Signal Control in a real-world simulation scenario." arXiv preprint arXiv:2103.16223.

Nabipour, N., et al. (2019). "Comparative analysis of machine learning models for prediction of remaining service life of flexible pavement." Mathematics **7**(12): 1198.

Naranjo-Pérez, J., et al. (2020). "A collaborative machine learning-optimization algorithm to improve the finite element model updating of civil engineering structures." Engineering Structures **225**: 111327.

Neves, A. C., et al. (2017). A new approach to damage detection in bridges using machine learning. International Conference on Experimental Vibration Analysis for Civil Engineering Structures, Springer.

Osorio-Lird, A., et al. (2018). "Application of Markov chains and Monte Carlo simulations for developing pavement performance models for urban network management." Structure and Infrastructure Engineering **14**(9): 1169-1181.

Patro, S. and K. K. Sahu (2015). "Normalization: A preprocessing stage." arXiv preprint arXiv:1503.06462.

Pellicer, E., et al. (2016). "Appraisal of infrastructure sustainability by graduate students using an active-learning method." Journal of Cleaner Production **113**: 884-896.

Peng, N., et al. (2021). "Urban Multiple Route Planning Model Using Dynamic Programming in Reinforcement Learning." IEEE Transactions on Intelligent Transportation Systems.

Piryonesi, S. M. and T. E. El-Diraby (2020). "Data analytics in asset management: Cost-effective prediction of the pavement condition index." Journal of Infrastructure Systems **26**(1): 04019036.

Qiao, J. Y., et al. (2021). "Policy implications of standalone timing versus holistic timing of infrastructure interventions: Findings based on pavement surface roughness." Transportation Research Part A: Policy and Practice **148**: 79-99.

Renard, S., et al. (2021). "Minimizing the global warming impact of pavement infrastructure through reinforcement learning." Resources, Conservation and Recycling **167**: 105240.

Sabour, M. R., et al. (2021). "Application of Artificial Intelligence Methods in Modeling Corrosion of Cement and Sulfur Concrete in Sewer Systems." Environmental Processes: 1-18.

Santero, N. J. and A. Horvath (2009). "Global warming potential of pavements." Environmental research letters **4**(3): 034011.

Santos, J., et al. (2020). "A fuzzy logic expert system for selecting optimal and sustainable life cycle maintenance and rehabilitation strategies for road pavements." International journal of pavement engineering: 1-13.

Schulman, J., et al. (2017). "Proximal policy optimization algorithms." arXiv preprint arXiv:1707.06347.

Song, Y., et al. (2021). "Developing sustainable road infrastructure performance indicators using a model-driven fuzzy spatial multi-criteria decision making method." Renewable and Sustainable Energy Reviews **138**: 110538.

Sutton, R. S. (1992). Introduction: The challenge of reinforcement learning. Reinforcement Learning, Springer**:** 1-3.

Tol, R. S. (2005). "The marginal damage costs of carbon dioxide emissions: an assessment of the uncertainties." Energy policy **33**(16): 2064-2074.

Torres-Machi, C., et al. (2017). "Towards a sustainable optimization of pavement maintenance programs under budgetary restrictions." Journal of Cleaner Production **148**: 90-102.

Van Otterlo, M. and M. Wiering (2012). Reinforcement learning and markov decision processes. Reinforcement learning, Springer**:** 3-42.

Visintine, B. A., et al. (2018). Validation of Pavement Performance Measures Using LTPP Data, United States. Federal Highway Administration. Office of Infrastructure ….

Walraven, E., et al. (2016). "Traffic flow optimization: A reinforcement learning approach." Engineering Applications of Artificial Intelligence **52**: 203-212.

Wang, F., et al. (2020). "A day-ahead PV power forecasting method based on LSTM-RNN model and time correlation modification under partial daily pattern prediction framework." Energy Conversion and Management **212**: 112766.

Wang, H. and R. Gangaram (2014). Life Cycle Assessment of Asphalt Pavement Maintenance, Rutgers University. Center for Advanced Infrastructure and Transportation.

Watts, N., et al. (2018). "The 2018 report of the Lancet Countdown on health and climate change: shaping the health of nations for centuries to come." The Lancet **392**(10163): 2479-2514.

Yamany, M. S., et al. (2020). "Characterizing the performance of interstate flexible pavements using artificial neural networks and random parameters regression." Journal of Infrastructure Systems **26**(2): 04020010.

Yao, L., et al. (2019). "Establishment of prediction models of asphalt pavement performance based on a novel data calibration method and neural network." Transportation Research Record **2673**(1): 66-82.

Yao, L., et al. (2020). "Deep reinforcement learning for long-term pavement maintenance planning." Computer-Aided Civil and Infrastructure Engineering **35**(11): 1230-1245.

Yu, B., et al. (2015). "Multi-objective optimization for asphalt pavement maintenance plans at project level: Integrating performance, cost and environment." Transportation Research Part D: Transport and Environment **41**: 64-74.

Zhang, B., et al. (2015). "Geometric reinforcement learning for path planning of UAVs." Journal of Intelligent & Robotic Systems **77**(2): 391-409.

Zhu, L., et al. (2021). "Operational Characteristics of Mixed-Autonomy Traffic Flow on the Freeway With On-and Off-Ramps and Weaving Sections: An RL-Based Approach." IEEE Transactions on Intelligent Transportation Systems.

Zolfpour-Arokhlo, M., et al. (2014). "Modeling of route planning system based on Q value-based dynamic programming with multi-agent reinforcement learning algorithms." Engineering Applications of Artificial Intelligence **29**: 163-177.

Huang, Y., et al. (2009). "Development of a life cycle assessment tool for construction and maintenance of asphalt pavements." Journal of Cleaner Production **17**(2): 283-296.

Keoleian, G. A. and T. A. Volk (2005). "Renewable energy from willow biomass crops: life cycle energy, environmental and economic performance." BPTS **24**(5-6): 385-406.

Labi, S. and K. C. Sinha (2005). "Life-cycle evaluation of flexible pavement preventive maintenance." Journal of transportation engineering **131**(10): 744-751.

Mandapaka, V., et al. (2012). "Mechanistic-empirical and life-cycle cost analysis for optimizing flexible pavement maintenance and rehabilitation." Journal of transportation engineering **138**(5): 625-633.

Praticò, F., et al. (2011). "Comprehensive life-cycle cost analysis for selection of stabilization alternatives for better performance of low-volume roads." Transportation Research Record **2204**(1): 120-129.

Santero, N. J. and A. Horvath (2009). "Global warming potential of pavements." Environmental research letters **4**(3): 034011.

Sutton, R. S. (1992). Introduction: The challenge of reinforcement learning. Reinforcement Learning, Springer**:** 1-3.

Tol, R. S. (2005). "The marginal damage costs of carbon dioxide emissions: an assessment of the uncertainties." Energy policy **33**(16): 2064-2074.

Visintine, B., et al. (2017). Validation of Pavement Performance Measures Using LTPP Data: Final Report, Report No. FHWA-HRT-17-089. Federal Highway Administration, Washington, DC.

Wang, T., et al. (2012). "Life cycle energy consumption and GHG emission from pavement rehabilitation with different rolling resistance." Journal of Cleaner Production **33**: 86-96.

Yao, L., et al. (2020). "Deep reinforcement learning for long-term pavement maintenance planning." Computer-Aided Civil and Infrastructure Engineering **35**(11): 1230-1245.

Khandel, O., & Soliman, M. (2019). Integrated framework for quantifying the effect of climate change on the risk of bridge failure due to floods and flood-induced scour. *Journal of bridge engineering*, *24*(9), 04019090.

Khandel, O., & Soliman, M. (2021a). Integrated framework for assessment of time-variant flood fragility of bridges using deep learning neural networks. *Journal of Infrastructure Systems*, *27*(1), 04020045.

Khandel, O., Soliman, M., Floyd, R. W., & Murray, C. D. (2021b). Performance assessment of prestressed concrete bridge girders using fiber optic sensors and artificial neural networks. *Structure and Infrastructure Engineering*, *17*(5), 605-619.

Sheikh, I. A., Khandel, O., Soliman, M., Haase, J. S., & Jaiswal, P. (2021). Seismic fragility analysis using nonlinear autoregressive neural networks with exogenous input. *Structure and Infrastructure Engineering*, 1-15.